\documentclass[letterpaper, 10 pt, conference]{ieeeconf}  

\IEEEoverridecommandlockouts 

\usepackage{graphics} 
\usepackage{amsmath,amssymb,amsfonts} 
\usepackage{bm}
\usepackage{subcaption}	 
\usepackage{graphicx}
\usepackage{multirow}
\usepackage{array}
\usepackage{url}
\usepackage{floatrow}

\usepackage{booktabs}
\usepackage[table,xcdraw]{xcolor}

\usepackage{custom_commands}
\usepackage{xcolor}

\usepackage{todonotes}

\usepackage[colorlinks = true,%
			linkcolor  = red,%
			anchorcolor= red,
			citecolor  = red
			]{hyperref}

\title{\LARGE \bf
CLINS: Continuous-Time Trajectory Estimation for \\ LiDAR-Inertial System
}

\author{Jiajun Lv$^{1,\dag}$, Kewei Hu$^{1,\dag}$, Jinhong Xu$^{1}$, Yong Liu$^{1}$, Xiushui Ma$^{2}$, Xingxing Zuo$^{1}$
\thanks{$^1$ April Lab, Zhejiang University, Hangzhou, China. (Yong Liu is the corresponding author, email: yongliu@iipc.zju.edu.cn)}%
\thanks{$^2$ NingboTech University, Ningbo, China.}
\thanks{$^\dag$ Jiajun Lv and Kewei Hu contribute equally to this work.}
}

\begin{document}

\maketitle
\setlength{\textfloatsep}{4.0pt} 

\begin{abstract}
In this paper, we propose a highly accurate continuous-time trajectory estimation framework dedicated to SLAM (Simultaneous Localization and Mapping) applications, which enables fuse high-frequency and asynchronous sensor data effectively.  We apply the proposed framework in a 3D LiDAR-inertial system for evaluations. The proposed method adopts a non-rigid registration method for continuous-time trajectory estimation and simultaneously removing the motion distortion in LiDAR scans. Additionally, we propose a two-state continuous-time trajectory correction method to efficiently and efficiently tackle the computationally-intractable global optimization problem when loop closure happens. 
We examine the accuracy of the proposed approach on several publicly available datasets and the data we collected.
The experimental results indicate that the proposed method outperforms the discrete-time methods regarding accuracy especially when aggressive motion occurs. Furthermore, we open source our code at \url{https://github.com/APRIL-ZJU/clins} to benefit research community.
\end{abstract}
\section{Introduction}
Multi-sensor fusion plays an essential role in simultaneous localization and mapping (SLAM) algorithms with its complementarity and robustness, and it has been widely deployed in autonomous navigation, scene reconstruction, mixed reality, etc.
In this paper, we study the high-accuracy continuous-time trajectory estimation with a fusion of LiDAR and IMU measurements.
Most existing methods process LiDAR and IMU measurements in a discrete-time fashion.  
The LiDAR scan collecting points when the laser heads rotate around a mechanical axis, thus motion distortion is unavoidable when the LiDAR does not keep static in the data collecting process. Discrete-time based methods undistort the LiDAR points into the start time instant of the LiDAR scan by interpolations. IMU measurements are interpolated and integrated to formulate relative poses constraints between discrete LiDAR scans.
%
Discrete-time based methods have several inherent limitations that hurt the estimation. Firstly, in practice, different sensors do not get measurements at the same frequency, let alone the same time instants. Interpolations has to be employed to fuse measurements from different sensors, which introduces non-negligible errors. 
Secondly, it is unlikely to leverage the raw measurements directly, i.e. the raw LiDAR points and raw IMU measurements. Raw LiDAR points are undistorted into the specific time instants to constitute LiDAR scans, while IMU measurements are assembled to get integrated relative pose measurements. These phenomenons result from the intractable super-high-frequency raw sensor measurements, which requires huge amount of pose variable to be estimated if they are utilized in a direct way.
The above-mentioned difficulties can be summarized as the discrete pose representation fails to meet the system's demand of high temporal resolution. 
Recently, the continuous-time based method, which models the trajectory as a function of time and supports querying poses at any timestamp that naturally solves the problem of integrating asynchronous and high-frequency data. With those sound properties, continuous-time based method has been applied to many areas, such as visual-inertial navigation system~\cite{lovegrove2013spline, ovren2019trajectory}, event camera~\cite{mueggler2018continuous}, rolling-shutter camera~\cite{lovegrove2013spline}, actuated LiDAR~\cite{alismail2014continuous}, intrinsic and extrinsic calibration between sensors~\cite{furgale2012continuous,lv2020targetless}.
 
\begin{figure}[t]
	\centering
	\includegraphics[width=1.0\columnwidth]{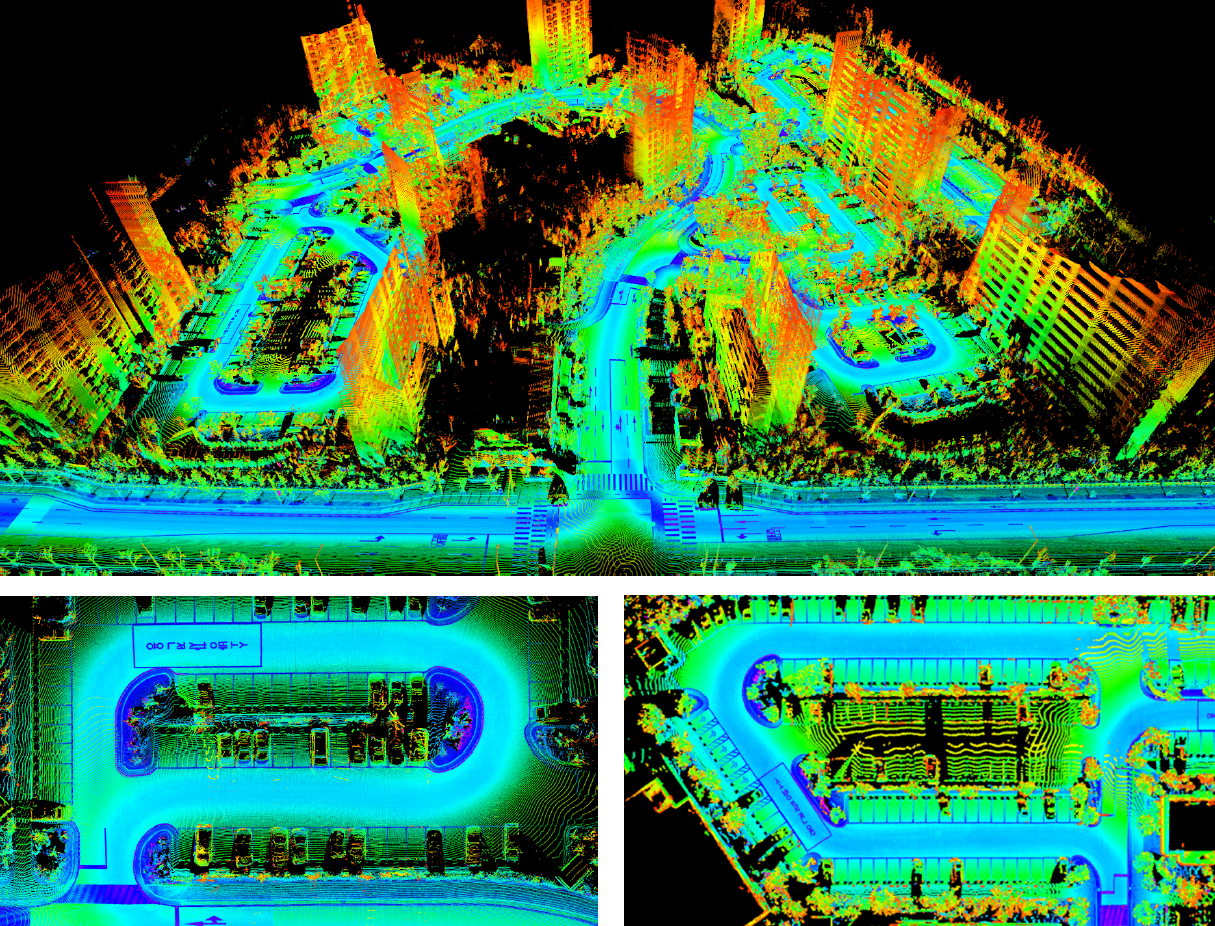}
	\caption{Dense 3D reconstruction of Kaist-Urban-07 dataset by simply assembling 2D LiDAR scans from SICK LMS-511 with the estimated continuous-time trajectory from CLINS. The global trajectory is estimated with 3D LiDAR, Velodyne VLP-16 and Xsens IMU, MTi-300. 2D LiDAR scans is accumulated for reconstruction due to its high density.}
    \label{fig:kaist_map}
    \vspace{-0.5em}
\end{figure}

This paper proposes a complete continuous-time trajectory estimation framework for LiDAR-inertial system.
We summarize the contributions as follows:
\begin{itemize}
    \item We propose a continuous-time trajectory estimator, which now supports the fusion of 3D LiDAR points and inertial data, and it is easy to expand and fuse data from other asynchronous sensors at arbitrary frequencies.
    \item We propose a two-stage continuous-time trajectory correction method to efficiently and effectively tackle loop closures.
    \item The proposed approach is extensively evaluated on several publicly available datasets and our collected datasets, and compared to several state-of-the-art methods. We further make the code open-sourced. To the best of our knowledge, this is the first open-sourced continuous-time LiDAR-inertial trajectory estimator.
\end{itemize}
\section{Related Works}

Continuous-time state estimation for solving SLAM problem is firstly systematically derived in ~\cite{furgale2012continuous} by Furgale et al. They firstly employ the continuous-time batch optimization on calibrating the rigid transformation between camera and IMU in visual-inertial system, and successively explore to calibrate  temporal offset between the camera and IMU~\cite{furgale2013unified}, shutter timings for rolling-shutter camera~\cite{oth2013rolling}, spatio-temporal extrinsics between a LiDAR and a stereo-visual-inertial system~\cite{rehder2014spatio}. Rehder et al. summarize their works and develop a general framework~ \cite{rehder2016general} for general spatio-temporal calibration between diverse sensors and evaluate on several combinations of different sensors in support of the generality claim. In addition to the calibration applications of continuous-time state estimation, Furgale et al. also present the details of theoretical derivations and effective implementation using B-splines in the well-known Kalibr calibration toolbox~\cite{furgale2015continuous}. Sommer et al. further investigate the initialization and analytical Jacobians of B-splines on Lie group~\cite{furgale2015continuous}.
This series of works has made a significant contribution to the continuous-time state estimation.

Recently, continuous-time trajectory method has been employed in LiDAR odometry. In ~\cite{zlot2014efficient, kaul2016continuous}, researchers propose a continuous SLAM solution with a spinning 2D laser scanner which is relatively dense and friendly for surfel-based registration.
Alismail et al. propose continuous-ICP~\cite{alismail2014continuous} that explicitly accounts for sensor motion during registration which improves the accuracy compared to rigid registration. 
Since global batch optimization of continuous-time trajectory is highly time-consuming, Droeschel et al. ~\cite{droeschel2018efficient} present a hierarchical refinement structure that optimizes a single firing sequence based sub-graph firstly and incorporate sub-graph when optimizing the allocentric pose graph.
Park et al. ~\cite{park2018elastic} adopt a map-centric method which introduces map deformation to remove the need for global trajectory optimization. 
Instead of adjusting the global continuous trajectory directly when loop closure happens, we propose a two-state based continuous trajectory correction method. We firstly perform a discrete-time pose graph optimization involved with key-scan poses; then the continuous-time trajectory is aligned with the optimized key-scan poses while maintaining local shape via original local velocities.

\section{Representation of Continuous-time Trajectory}
First of all, we introduce notations used in this paper. We denote the 6-DoF rigid transformation by ${}^B_A\mat{T}\in \SE(3) \in \R^{4\times4}$, which transforms the point ${}^A\vect{p}\in \R^3$ in the frame $\{A\}$ into frame $\{B\}$. ${}^B_A\mat{T} = \begin{bmatrix} {}^B_A\mat{R} &  {}^B\vect{p}_A \\ \mathbf{0} & 1\end{bmatrix}$ consists of rotational part ${}^B_A\mat{R}\in \SO(3)$ and translational part ${}^B\vect{p}_A\in \R^3$. 
For simplicity, we omit the homogeneous conversion in the rigid transformation by $ {}^B\vect{p} = {}^B_A\mat{T}  {}^A\vect{p}$.   
$\Exp(\cdot)$ maps tangent vector in $\R^3$ to Lie group $\SO(3)$, and $\Log(\cdot)$ is its inverse operation. $(\cdot)_\vee$ is used to map the elements in the Lie algebra to tangent vectors.

\begin{table}[t]
\resizebox{\linewidth}{!} {
\renewcommand{\arraystretch}{1.4} 
	\centering
    \caption{Summarization of uniform B-Spline based continuous-time trajectory representation.}
    \label{table:traj_representation}
    \begin{tabular}{|c|c|}
        \toprule 
        \multicolumn{2}{|c|} {\textbf{Uniform B-Spline Basics}} \\
        \midrule  
        Order &  $k$  \\
        Knot distance &  $\Delta t$  \\  
        Coeff. Vec. $(^1)$ 
        & $\vect{\phi}(u) = \mat{{M}}^{(k)} \vect{u}$ \\
        Cumul. Coeff. Vec.$(^{1,2})$ 
        & $\vect{\lambda}(u) = \mat{\widetilde{M}}^{(k)} \vect{u}$ \\
        \hline  
        \hline  
         \multicolumn{2}{|c|} {\textbf{Position in Cumulative Form}} \\
        \midrule  
        Control point & $\mat{\Phi}_p = \{\vect{p}_i\} \in\R^3, i\in[0,n]$  \\
        Distance 
        & $\vect{d}_j^i = \vect{p}_{i+j}-\vect{p}_{i+j-1} \in \R^3$  \\
        Position
        & $\vect{p}(u) = \vect{p}_i + \sum_{j=1}^{k-1}{\lambda_j(u)\cdot\vect{d}^i_j}$  \\
        Velocity  
        & $\vect{v}(u)=\sum_{j=1}^{k-1}{\dot{\lambda}_j(u)\cdot\vect{d}^i_j}$ \\
        Acceleration  & $\vect{a}(u)=\sum_{j=1}^{k-1}{\ddot{\lambda}_j(u)\cdot\vect{d}^i_j}$  \\
        \hline  
        \hline  
        \multicolumn{2}{|c|} {\textbf{Orientation in Cumulative Form}} \\
        \midrule  
        Control point & $\mat{\Phi}_R = \{\mat{R}_i\}\in\SO(3),i\in[0,n]$  \\
        Distance 
        & $\vect{d}^i_j = \Log\left(\mat{R}_{i+j-1}^{-1}\mat{R}_{i+j}\right) \in \R^3$ \\
        Position
        & $\mat{R}(u) = \mat{R}_i \cdot \prod_{j=1}^{k-1}{\Exp\left(\lambda_j(u)\cdot\vect{d}^i_j\right)}$ \\
        Velocity  
        & $\vect{\omega}(u) = (\mat{R}^{\top} \dot{\mat{R}})_{\vee} $ \\
        \bottomrule 
        \multicolumn{2}{l}{$(^1)$ $t\in [t_i, t_{i+1})$,\quad $u(t) = s(t)-i, \quad  s(t) := (t - t_0)/\Delta t$. }\\
        \multicolumn{2}{l}{$(^2)$ Note that $\lambda_0(u)\equiv 1$.}
    \end{tabular}
}
\vspace{-1em}
\end{table}

B-spline is smooth ($C^2$ continuity in case of cubic spline) and has local continuity. Most importantly, it has closed-form analytic derivatives which is easy to match against IMU measurements. To this end, we adopt two separate groups of B-splines to parameterize the 3D translation and 3D rotation, $\vect{p}(t) \in \R^{3}$ and $\mat{R}(t) \in \SO(3)$. This formulation is termed as split representation of continuous-time trajectory in ~\cite{haarbach2018survey,ovren2019trajectory}. A comprehensive overview of the 
B-spline can be found in~\cite{sommer2020efficient}. Here we also summarize the definition of continuous trajectory representation by uniform B-splines in Tab.~\ref{table:traj_representation}.
Specifically, spline matrix $\mat{{M}}^{(k)}$ and cumulative matrix $\mat{\widetilde{M}}^{(k)}$ are constant for uniform B-spline~\cite{sommer2020efficient}. 
For translational trajectory representation in $\R^n$ , the basic form and the cumulative form of the B-spline representation are equivalent and can be converted to each other. However, this is not the case in the non-Euclidean space $\SO(3)$, The orientational trajectory representation is feasibly represented in the cumulative form. Tab.~\ref{table:traj_representation} lists cumulative form trajectory in $\R^3$ and $\SO(3)$ that are adopted in this paper.
Taking derivative of splines with respect to time, we can get velocity and acceleration. As shown in Tab.~\ref{table:traj_representation}, linear velocity $\vect{v}(t)$ and linear acceleration $\vect{a}(t)$ are in global frame while angular velocity $\vect{\omega}(t)$ is in local frame.

The continuous-time trajectory of IMU in global frame $\{G\}$ is denoted as 
\begin{align}
   {}^G_I\mat{T}(t) = \begin{bmatrix}\mat{R}(t)& \vect{p}(t) \\ \mathbf{0} & 1\end{bmatrix}\text{,}
\end{align}
where we omit the superscripts and subscripts for simplicity. 
With the known pre-calibrated extrinsic transformation ${}^I_L\mat{T}$ between IMU and LiDAR using toolbox LI-Calib~\cite{lv2020targetless}, the trajectory of LiDAR could be calculated as \begin{align}
    {}^G_L\mat{T}(t) = {}^G_I\mat{T}(t) {}^I_L\mat{T}\text{.}
\end{align}
Due to the locality of the B-spline basis, for $t \in [t_i, t_{i+1})$, ${}^G_I\mat{T}(t)$ is only controlled by knots at $ \{t_i,t_{i+1}, \cdots, t_{i+k-1}\} $ with their corresponding control points set $\mat{\Phi}(t_i,t_{i+1})$:
\begin{equation}
\begin{split}
    \mat{\Phi}(t_i,t_{i+1}) &= \mat{\Phi}_R(t_i,t_{i+1}) \cup \mat{\Phi}_p(t_i,t_{i+1}) \\
    &=\{ \mat{R}_i,\cdots,\mat{R}_{i+k-1}\} \cup \{\vect{p}_i,\cdots,\vect{p}_{i+k-1}\}   \text{.}
\end{split}
\end{equation}

\section{Methodology}

\begin{figure}[t]
	\centering
	\includegraphics[width=1.0\columnwidth]{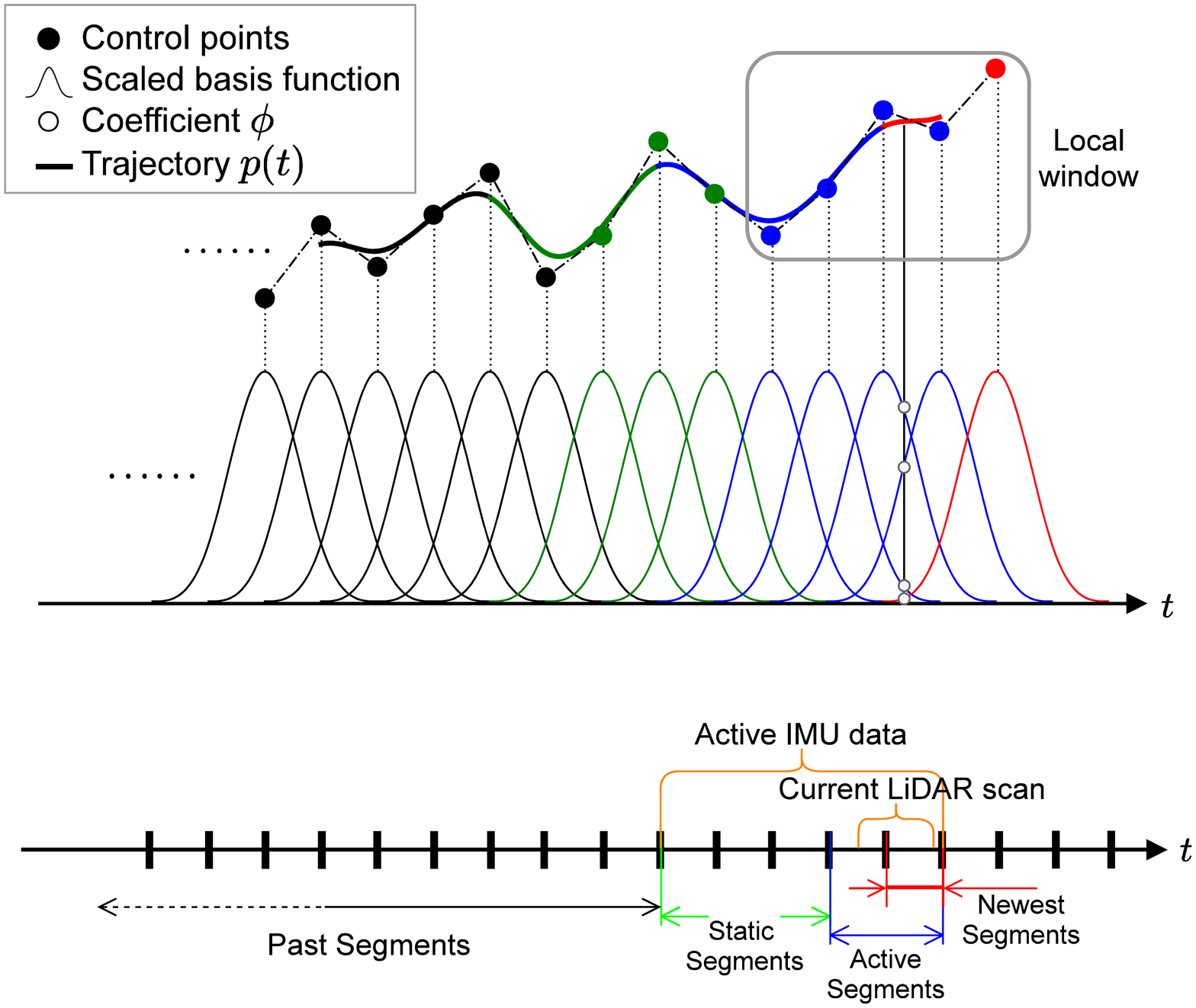}
	\caption{An illustration of the proposed continuous-time trajectory for LiDAR-inertial system based on cubic B-splines. }
	\label{fig:overview}
	\vspace{-0.5em}
\end{figure}

Fig.~\ref{fig:overview} illustrates the definitions involved in the proposed continuous-time trajectory estimator of LiDAR-inertial system. We define the \textit{active segments} of time based on the data-collection time interval of current LiDAR scan. The control point sequence $\mat{\Phi}_{\text{active}}$ corresponding to the active segments are denoted as \textit{active control points} (blue dots and red dots). The subset of basis functions corresponding to $\mat{\Phi}_{\text{active}}$ are denoted as active basis functions (blue curves and red curves). The active trajectory (in blue and red) is determined by the active control points in a local region. Furthermore, except the active segments of time, the time segments involved with the active basis functions are notated as \textit{static segments}. Accordingly, \textit{static control points} $\mat{\Phi}_{\text{static}}$ (green dots) and static basis functions (green curves) are defined.

In the proposed system, after the coming of a new LiDAR scan, new control points (red dots) are added into the state vector to parameterize the extended trajectory.  We utilize the integrated IMU poses to initialize these new control points (see Sec.~\ref{sec:initial}). 
Subsequently, we extract LOAM features~\cite{zhang2014loam} including the edge points and planar points from current scan to make data association with local submap, which is composed of plenty of selected key-scans based on spatio-temporal distance.  
Finally, given the raw IMU measurements in the local window and the associated LiDAR features, we estimate the continuous-time trajectory in a batch optimization fashion.
The state estimation problem could be formulated as a maximum a posteriori (MAP) problem. With the assumption of independent Gaussian noise corruption on the raw sensor measurements, we can estimate the continuous-time trajectory by solving a non-linear least squares (NLLS) problem.

\subsection{Initialization} \label{sec:initial}
When new LiDAR scan arrives, new control points are added to extend the existing  trajectory. We integrate discrete IMU data to obtain high-frequency rotation, position and linear velocity estimations, which are denoted as $\mat{R}_{I_m}, \vect{p}_{I_m}, \vect{v}_{I_m}$ at time $t_m$, respectively.
We can minimize the following cost function to initialize the newly added control points $\mat{\Phi}_\text{new}$
\begin{equation}
\begin{split}
     \arg\min_{\mat{\Phi}_\text{new}} \sum \Big(
    \norm{\Log(\mat{R}^{\top}_{I_m} \mat{R}(t_m))} + \norm{\vect{p}(t_m) - \vect{p}_{I_m}} + \\
    \norm{\vect{v}(t_m) - \vect{v}_{I_m}} \Big)\text{.}
\end{split}
\end{equation}


\subsection{Non-rigid Registration in Local Window}
For new-coming scan $\mathcal{S}_k$, measuring during time interval $[t_k,t_k+\Delta T]$, where $t_k$ is the timestamp of the first point in scan $\mathcal{S}_k$ and $\Delta T$ is the period of completing a LiDAR scan, it is essentially a non-rigid point cloud in  $\mathcal{S}_k$ if there is external motion while sensor is scanning. Therefore traditional registration algorithms, which compute the relative transformation between two rigid point clouds, are not applicable or have degraded performance. To tackle this problem, we propose a non-rigid registration method which estimates the continuous-time trajectory in current scan. 

Specifically, each point in $\mathcal{S}_k$ is transformed to a unified frame $\{L_k\}$ 
\begin{align}
    {}^{L_k}\vect{x}_{kj} = {}^G_L\mat{T}(t_k)^{\top}{}^G_L\mat{T}(t_k+\tau_j) {}^{L_{kj}}\vect{x}_{kj}
\end{align}
where $\tau_j$ is relative timestamp of point ${}^{L_{kj}}\vect{x}_{kj}$ with respect to the start time of current scan. 
Similar to ~\cite{zhang2014loam}, in order to keep the computation tractable, we extract the planar and edge features from the raw point cloud by computing local curvature. These extracted features are used in the following registration.

Local submap consists of some selected key-scans that are distributed in time or space. Note that when the changes of system’ pose are over a certain extent, a key-scan is selected and all the points in that key-scan are undistorted into the start of the LiDAR scan by using the non-active trajectory. To find current features' correspondences in the local submap, we transform current features into the frame of local submap and make correspondences based on closet neighbors~\cite{zhang2014loam}.

Considering LiDAR data degenerate in some structureless environments while the high-frequency IMU data are not affected, we tightly couple IMU measurements with LiDAR features to constrain the trajectory. 
Thus the non-rigid registration problem can be defined as: given the associated LiDAR features in current scan and inertial measurements in active segments and static segments, estimate the active control points $\mat{\Phi}(t_k,t_k+\Delta T)$ of the trajectory and the biases of IMU. Specifically, we can solve this problem by minimizing the following objective function
\begin{align}
\label{eq:opt_problem}
    \arg\min_{\mathcal{X}}\sum \norm{\vect{r}_{\mathcal{L}}}_{\mat{\Sigma}_\mathcal{L}} + \norm{\vect{r}_a}_{\mat{\Sigma}_a} +
    \norm{\vect{r}_w}_{\mat{\Sigma}_w}
\end{align}
where $\mathcal{X} = \{\mat{\Phi}(t_k,t_k+\Delta T), \vect{b}_a, \vect{b}_g\}$, and $\vect{b}_a$ , $\vect{b}_w$ are  the bias of accelerator and gyroscope, respectively. $\vect{r}_{\mathcal{L}},\vect{r}_a,\vect{r}_w$ are residual errors associated to LiDAR features and IMU measurements, respectively. $\mat{\Sigma}_\mathcal{L},\mat{\Sigma}_a, \mat{\Sigma}_w$ are the corresponding covariance matrices. The residuals are defined as
\begin{align}
    \vect{r}_{\mathcal{L}} &=  \pi\left({}^G_L\mat{T}(t_k+\tau_j)   {}^{L_{kj}}\vect{x}_{kj}\right) \\
    \vect{r}_{a} &= {}^G_I\mat{R}^{\top}(t_m)\left(\vect{a}(t_m)-{}^G\vect{g}\right) - \vect{a}_m + \vect{b}_a\\
    \vect{r}_{\omega} &= \vect{\omega}(t_m) - \vect{\omega}_m + \vect{b}_w
\end{align}
where project function $\pi(\cdot)$ is a point-to-plane projection for planar features and a point-to-line projection for edge features. $\vect{a}_m, \vect{\omega}_m$ are inertial measurements at time $t_m$. Note that static control points are involved in the optimization process but remain unchanged during the optimization. We use the Levenberg–Marquardt method implemented in Ceres Solver~\cite{ceres-solver} to solve the above non-linear problem.

\begin{figure}[t]
	\centering
	\includegraphics[width=0.8\columnwidth]{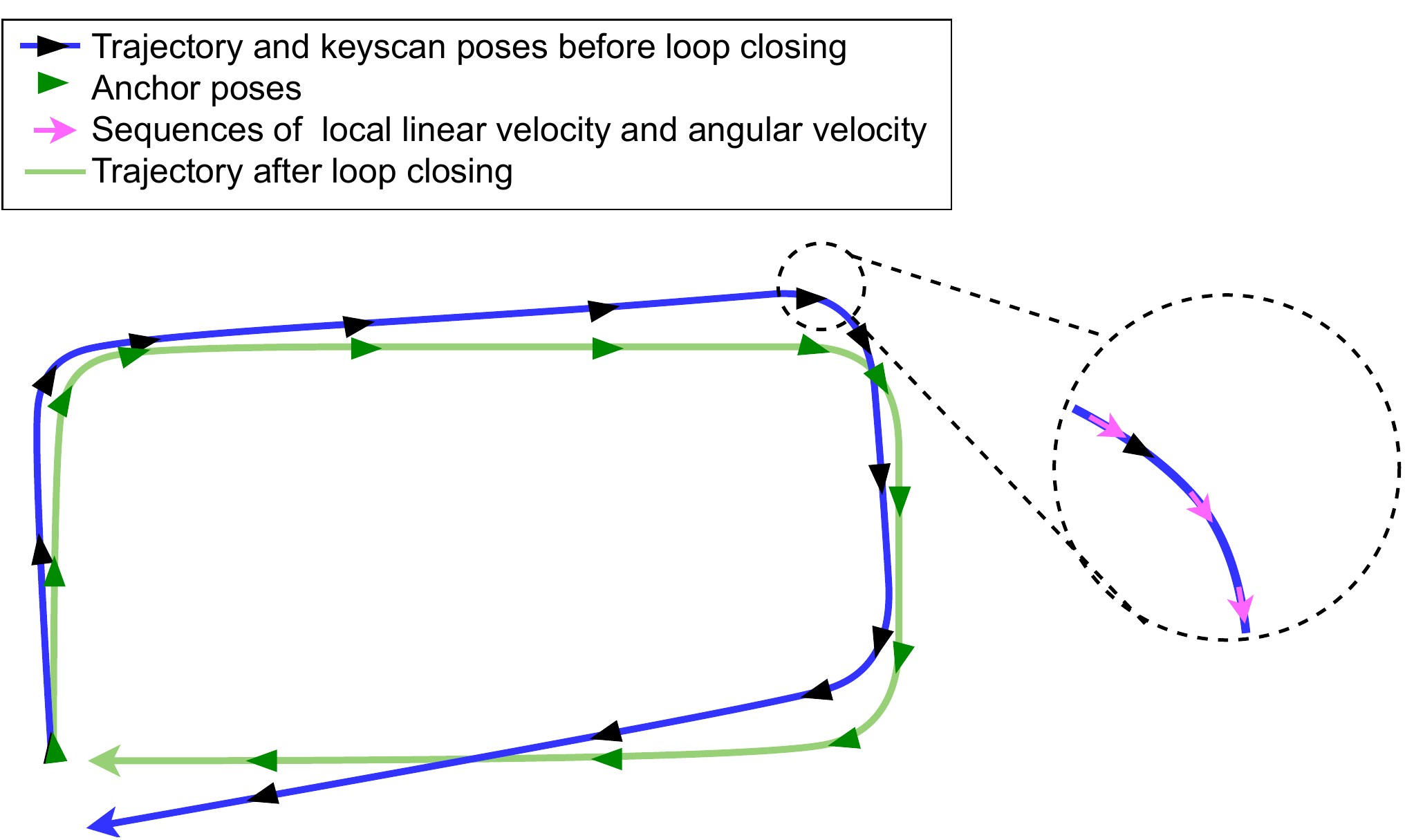}
	\caption{An illustration of the two-stage continuous-time trajectory correction method for loop closures.}
	\label{fig:loop_closing}
	\vspace{-0.5em}
\end{figure}

\subsection{Trajectory Correction}
Accumulative estimation drift is unavoidable in odometry systems, we also correct the estimations when loop closure occurs to mitigate the drift. Normallly, global optimization is required for smooth and consistent trajectory correction.
Considering the fact that it is really time-consuming to perform global optimization on the continuous trajectory with abundant of knots, we propose a two-stage trajectory correction method that is effictive but computationally friendly. At stage one, when loop closures are detected, we perform a pose-graph optimization over the involved discrete poses of key-scans to eliminate cumulative draft. This stage is same with the traditional descrete-time based method~\cite{Mur-Artal2015}. At stage two, we try to update the control points with the updated poses of key-scans. This operation can be interpreted as projecting the continuous trajectory to some anchored poses and preserving the local shape of the trajectory. In general, the estimation results of the inertial-aided LiDAR odometry is accurate in local regions, so we compute the local linear and angular velocities from the original continuous-time trajectory at certain time instants. Those velocities can be used to locally constrain the shape of the trajectory. To this end, the potential problem at state two to be solved could be explicitly formulated as
\begin{equation}
\begin{split}
    \arg\min_{\mat{\Phi}_{\text{update}}}\sum \left(
    \norm{\Log(\hat{\mat{R}}^{\top}_k \mat{R}(t_k))} + 
    \norm{\vect{p}(t_k) - \hat{\vect{p}}_k} \right)+ \\
    \sum \left(
    \norm{\mat{R}(t_j)^{\top}\vect{v}(t_j) - \hat{\vect{v}}_j} +
    \norm{\vect{\omega}(t_j) - \hat{\vect{\omega}}_j} \right)
\end{split}
\label{eq:loop_problem}
\end{equation}
where $\hat{\mat{R}}_k, \hat{\vect{p}}_k$ are the updated poses of key-scan at time $t_k$ after the pose-graph optimization. And $\hat{\vect{v}}_j,\hat{\vect{\omega}}_j$ are the computed linear velocity and angular velocity at time $t_j$ from the trajectory before correction, respectively. Fig.~\ref{fig:loop_closing} illustrates the proposed two-stage based trajectory correction method.
\section{Experimental Results}

\begin{table}[t]
\caption{RMSE of translational and rotational estimation in the 6 sequences with motion varies from fast to slow.}
\resizebox{\linewidth}{!} {
\begin{tabular}{|c|c|c|c|c|c|}
\toprule
Error                                                                                 & Sequence & LOAM   & LIO-SAM & LIOM            & CLINS           \\ \hline
\multirow{6}{*}{\begin{tabular}[c]{@{}c@{}}Trans-\\ lation\\ RMSE\\ (m)\end{tabular}} 
& \cellcolor[HTML]{EFEFEF}fast1    & \cellcolor[HTML]{EFEFEF}0.4469 & \cellcolor[HTML]{EFEFEF}0.1058  & \cellcolor[HTML]{EFEFEF}0.0529          & \cellcolor[HTML]{EFEFEF}\textbf{0.0436} \\
& fast2    & 0.2023 & 0.1557  & 0.0663          & \textbf{0.0616} \\
& \cellcolor[HTML]{EFEFEF}mid1     & \cellcolor[HTML]{EFEFEF}0.1740 & \cellcolor[HTML]{EFEFEF}0.1486  & \cellcolor[HTML]{EFEFEF}0.0576          & \cellcolor[HTML]{EFEFEF}\textbf{0.0488} \\
& mid2     & 0.1010 & 0.0952  & 0.0874          & \textbf{0.0731} \\
& \cellcolor[HTML]{EFEFEF}slow1    & \cellcolor[HTML]{EFEFEF}0.0606 & \cellcolor[HTML]{EFEFEF}0.0727  & \cellcolor[HTML]{EFEFEF}0.0318          & \cellcolor[HTML]{EFEFEF}\textbf{0.0295} \\
& slow2    & 0.0666 & 0.0674  & 0.0435          & \textbf{0.0376} \\ \midrule
\multirow{6}{*}{\begin{tabular}[c]{@{}c@{}}Rotation\\ RMSE\\ (rad)\end{tabular}}      
& \cellcolor[HTML]{EFEFEF}fast1    & \cellcolor[HTML]{EFEFEF}0.1104 & \cellcolor[HTML]{EFEFEF}0.1047  & \cellcolor[HTML]{EFEFEF}\textbf{0.0537} & \cellcolor[HTML]{EFEFEF}0.0565          \\
& fast2    & 0.0763 & 0.1022  & 0.0574          & \textbf{0.0538} \\
& \cellcolor[HTML]{EFEFEF}mid1     & \cellcolor[HTML]{EFEFEF}0.0724 & \cellcolor[HTML]{EFEFEF}0.1020  & \cellcolor[HTML]{EFEFEF}\textbf{0.0523} & \cellcolor[HTML]{EFEFEF}0.0567          \\
& mid2     & 0.0617 & 0.0789  & 0.0567          & \textbf{0.0538} \\
& \cellcolor[HTML]{EFEFEF}slow1    & \cellcolor[HTML]{EFEFEF}0.0558 & \cellcolor[HTML]{EFEFEF}0.0698  &\cellcolor[HTML]{EFEFEF} 0.0496          & \cellcolor[HTML]{EFEFEF}\textbf{0.0438} \\
& slow2    & 0.0614 & 0.0715  & \textbf{0.0530} & 0.0570          \\ \bottomrule
\end{tabular}
}
\label{tab:vicon}
\end{table}


\begin{figure}[t]
	\centering
	\includegraphics[width=\columnwidth]{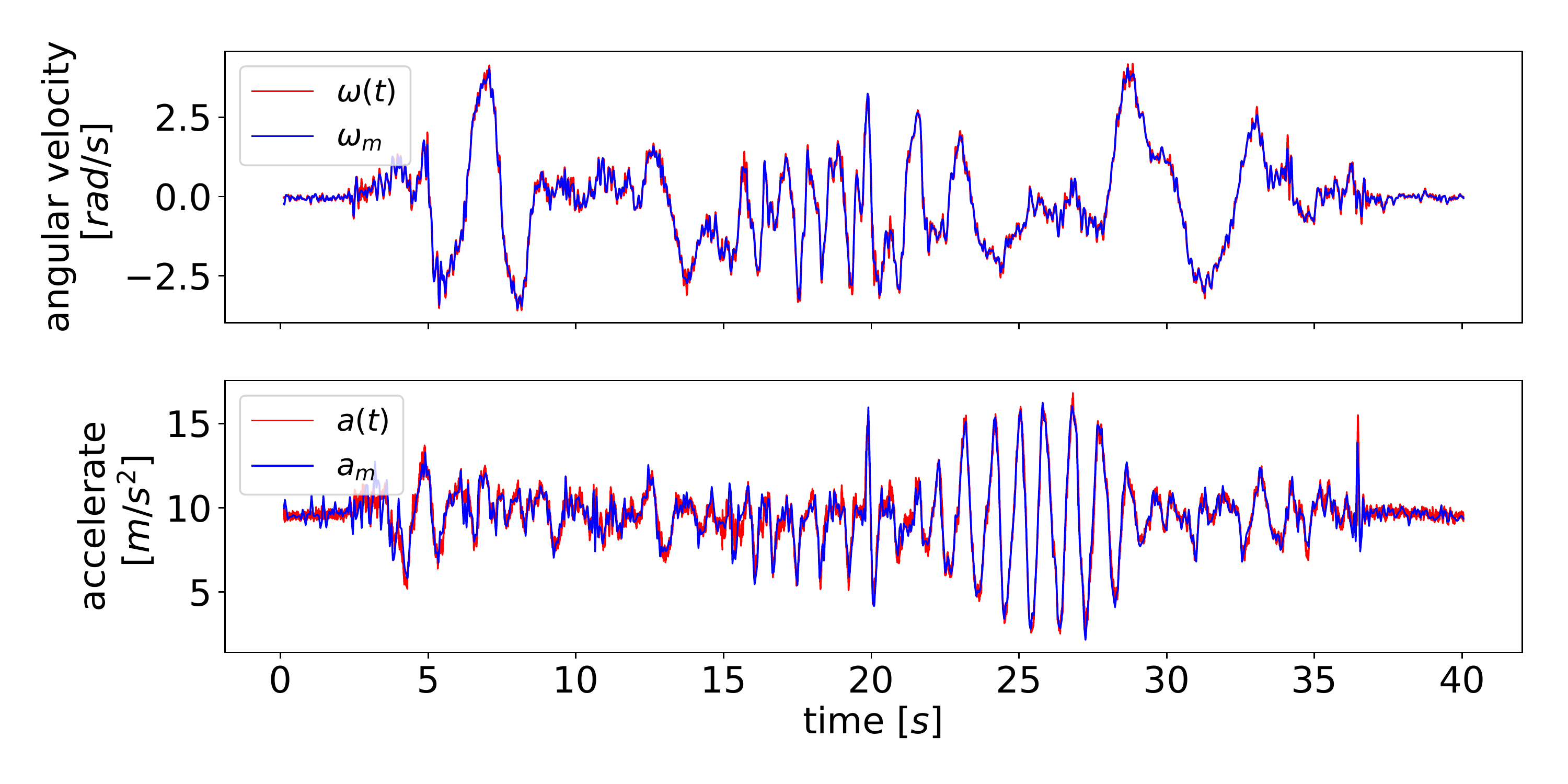}
	\vspace{-1em}
	\caption{The linear acceleration and angular velocity fitting results on \textit{fast1} sequence. Only the z-axis components are shown. Red is from the derivatives of the estimated continuous-time trajectory, while blue is from the raw IMU measurements.}
	\label{fig:fast1_seq}
	\vspace{-0.5em}
\end{figure}

\begin{figure}[t]
 	\centering
 	\includegraphics[width=0.4\columnwidth]{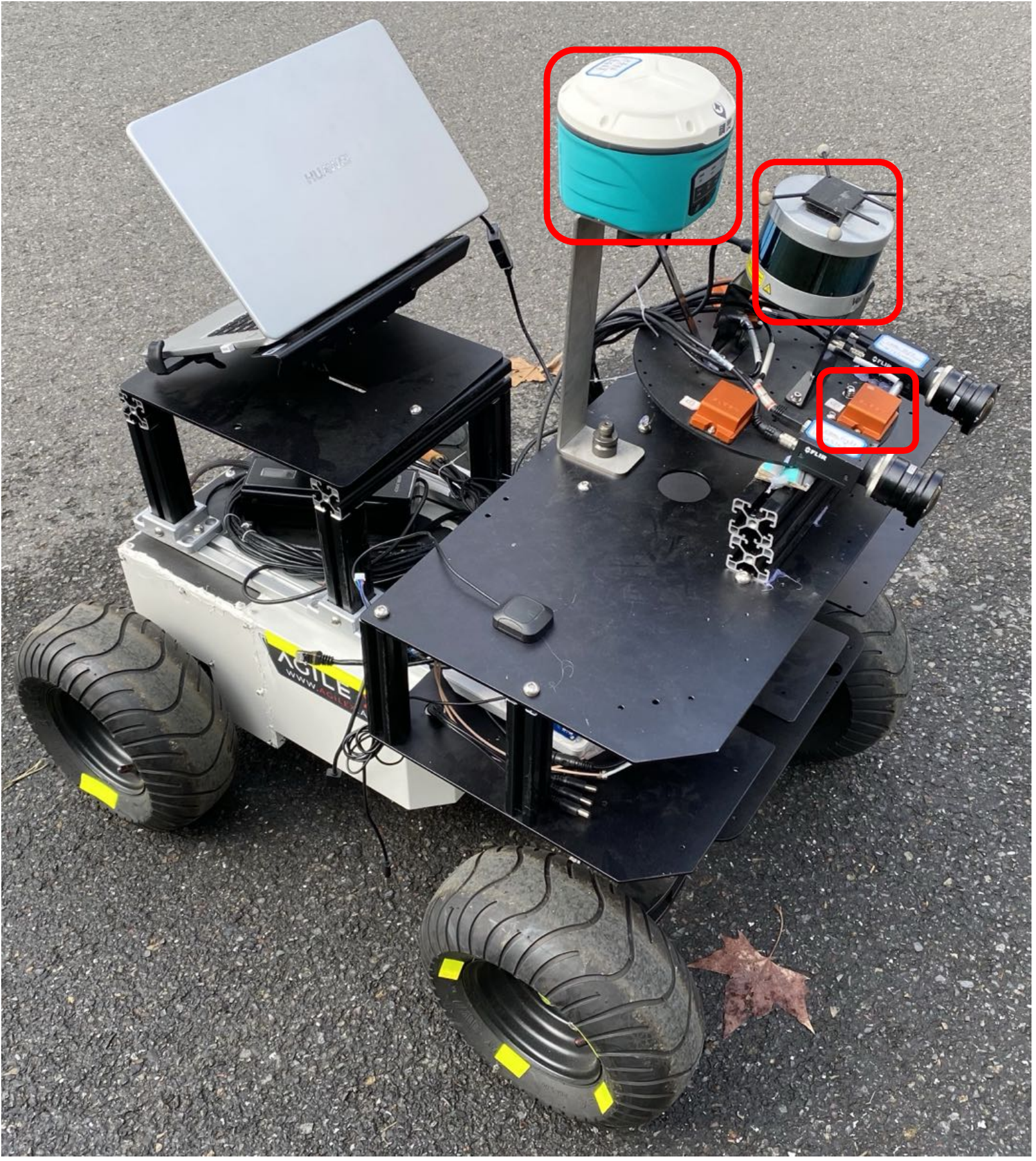}
 	\caption{The unmanned ground vehicle with self-assembled sensors rigidly mounted. Sensors with red box are used to collect YQ sequences in campus.}
 	\label{fig:car}
     \vspace{-0.5em}
\end{figure}

\begin{table}[t]
\resizebox{\linewidth}{!} {
\begin{tabular}{cccc}
\toprule
\multirow{2}{*}{Sequence} & Distance & Duration & \begin{tabular}[c]{@{}c@{}}Average Velocity\end{tabular} \\
                          & [km]     & [s]      & [m/s]                                                       \\ \midrule
YQ-01                     & 3.26     & 2471     & 1.32                                                        \\
YQ-02                     & 0.95     & 690      & 1.38                                                        \\
Kaist-Urban-07            & 2.54     & 570      & 4.46                                                        \\
Kaist-Urban-08            & 1.56     & 307      & 5.08                                 \\ \bottomrule
\end{tabular}
\caption{Some specifications of YQ and Kaist-Urban sequences.}
\label{tab:sequence_desci}
}
\end{table}

\begin{table}[t]
\resizebox{\linewidth}{!} {
\begin{tabular}{|c|cccc|}
\toprule
Method         & YQ-01          & YQ-02          & \begin{tabular}[c]{@{}c@{}}Kaist- \\Urban-07\end{tabular}  & \begin{tabular}[c]{@{}c@{}}Kaist- \\Urban-08\end{tabular}    \\ \midrule
\rowcolor[HTML]{EFEFEF} 
LIO-SAM (odom) & 8.857          & 3.215          & 1.288          & 3.524          \\ \hline
CLINS(odom)    & 5.917          & 3.15           & 1.383          & 3.907          \\ \hline
\rowcolor[HTML]{EFEFEF} 
LIOM           & 3.931          & \textbf{0.881}          & 1.515          & 16.277         \\ \hline
LIO-SAM        & \textbf{2.220}           & 2.487          & 1.972          & 3.951           \\ \hline
\rowcolor[HTML]{EFEFEF} 
CLINS          & 2.311 & 1.509 & \textbf{0.562} & \textbf{1.133} \\ \bottomrule
\end{tabular}
}
\caption{Summary of RMSE(m) of APE results on YQ and Kaist-Urban sequences. The best results are shown in bold.}
\label{tab:accuracy_summary}
\end{table}

\begin{figure}[t]
	\centering
	\begin{subfigure}{1.0\columnwidth} 
	    \centering
		\includegraphics[width=\columnwidth]{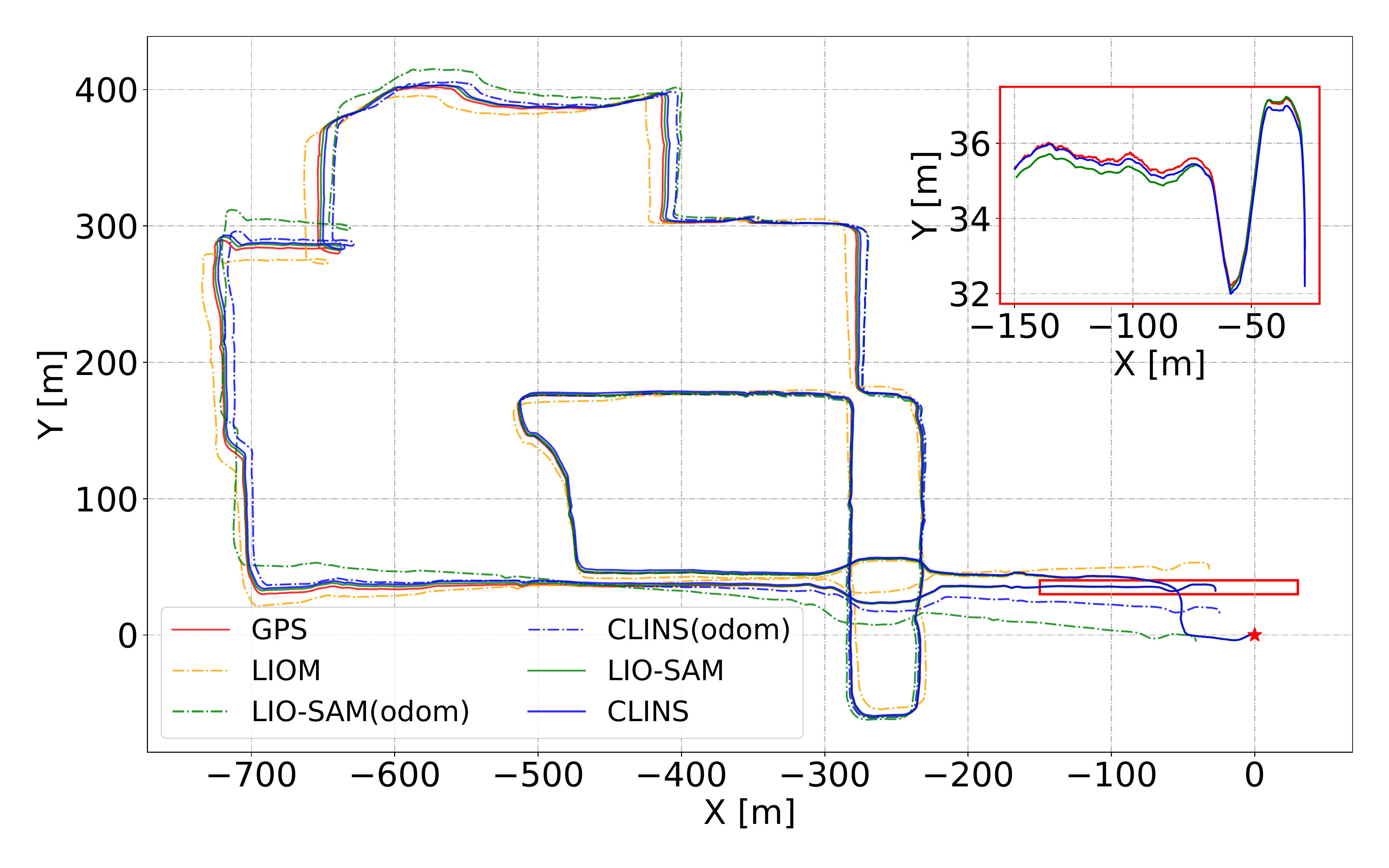}
	\end{subfigure}
	\begin{subfigure}{0.9\columnwidth}
	    \centering
		\includegraphics[width=\columnwidth]{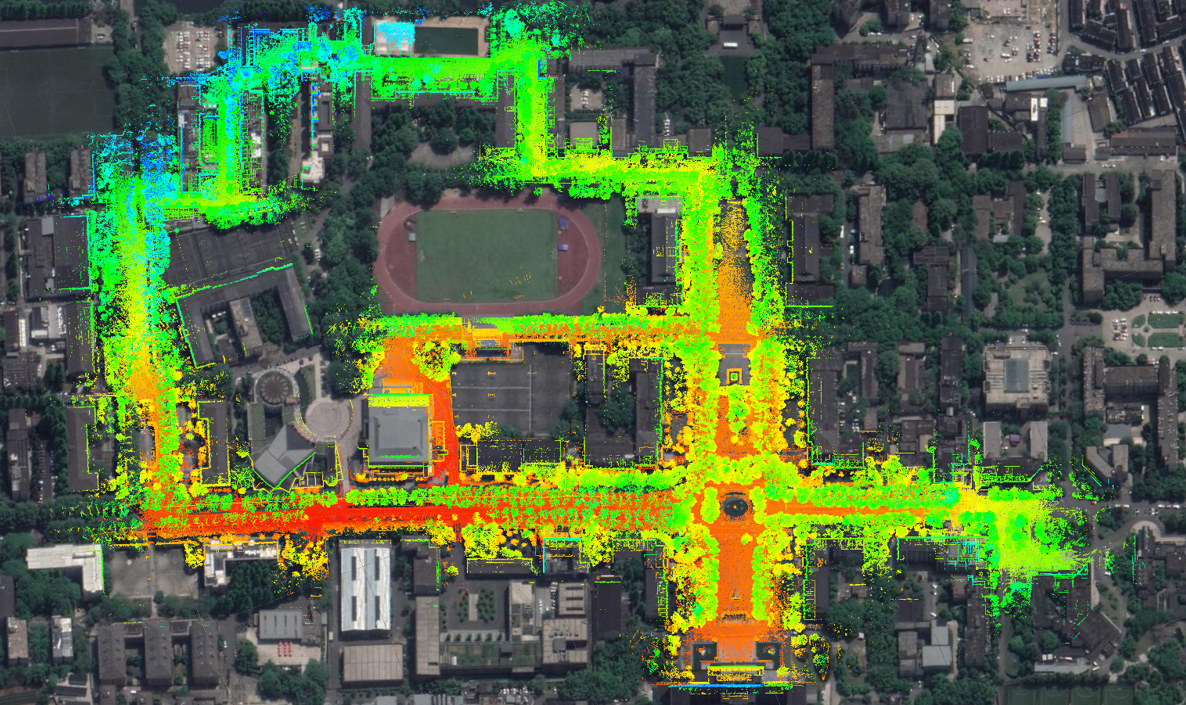}
	\end{subfigure}
 	\caption{Top: Trajectory comparison with different methods on YQ-01 sequence. The red star indicates the start position, and the end part of the trajectory are shown in zoom view. Bottom: Mapping results of CLINS with loop correction using YQ-01 Sequence. The map is consistent with the Google Earth imagery. }
 	\label{fig:map_of_yq01}
 	\vspace{-0.5em}
\end{figure}

\begin{figure*}[bt]
 	\centering
 	\includegraphics[width=0.95\columnwidth]{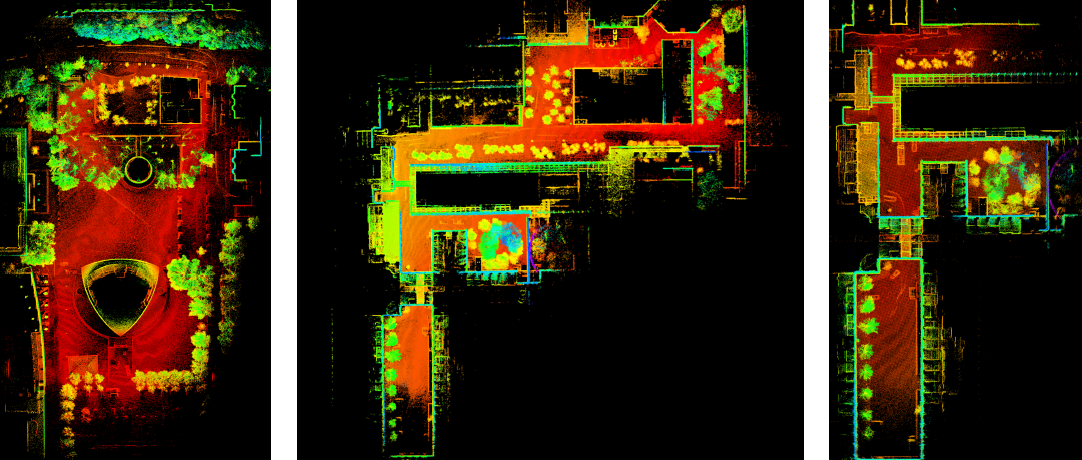}
 	\caption{Mapping results of CLINS using the \textit{garden}, \textit{walking} and \textit{small campus} datasets from left to right, respectively. All are colored with reflective intensity.}
 	\label{fig:mit_map}
 	\vspace{-0.5em}
\end{figure*}
	
We qualitatively and quantitatively evaluate the proposed method in a series of publicly available datasets as well as the datasets we collected.
Considering that there is no open-source continuous-time trajectory estimation method for LiDAR inertial system, we compare the proposed method with LIO-SAM~\cite{shan2020lio} and LIOM~\cite{ye2019tightly} (abbreviation of LIO-mapping). In the following experiments, LIO-SAM without loop correction is notated as LIO-SAM(odom), and CLINS(odom) for CLINS without loop correction.

\subsection{Trajectory Estimation in Room-scale Scenes}

We evaluate the proposed CLINS on the publicly available datasets\footnote{Available at \url{https://drive.google.com/drive/folders/1dPy667dAnJy9wgXmlnRgQZxQF\_ESuve3} } provided in~\cite{ye2019tightly} with ground truth to examine the representation capability of our continuous-time trajectory formulation and the estimation accuracy of positions and orientations. The knot distance $\Delta t$ of B-spline is set as 0.05 second to deal with the highly-dynamic motion. Tab~\ref{tab:vicon} summarizes the root mean square error (RMSE) of the estimations from different methods. Note that the experimental results for LOAM~\cite{zhang2014loam} and LIOM in Tab.~\ref{tab:vicon} are acquired directly from the paper~\cite{ye2019tightly}.
It is also important to note that the output pose estimation of LIOM is around 5 Hz and LIO-SAM in 10 Hz, while for CLINS, we query and evaluate the estimated pose at 100Hz from the obtained continuous-time trajectory .
From Tab.~\ref{tab:vicon} we can see that CLINS provides more accurate translation estimation in all sequences while the accuracy of rotation is similar between LIOM and CLINS. Fig.~\ref{fig:fast1_seq} shows the fitting results of the estimated trajectory on \textit{fast1} sequence compared to the raw IMU measurements. in which the angular velocity varies between -4.87 rad and 6.18 rad, while the acceleration reaches a maximum of 5.8 m/s$^2$. The experimental results on \textit{fast1} sequence in Tab.~\ref{tab:vicon} and Fig.~\ref{fig:fast1_seq} show that our proposed method can not only achieve the high accuracy in trajectory estimation, but also well fits the derivatives of the trajectories to the inertial measurements.

\subsection{Trajectory Estimation and Scene Mapping in Large-scale Scenes}

\textbf{Vehicle platform.}
We further investigate the accuracy of the CLINS with or without loop correction on outdoor long-distance sequence, YQ-01, YQ-02, Kaist-Urban-07 and Kaist-Urban-08. The first two sequences are collected on campus by ourselves using a Velodyne VLP-16 LiDAR, an Xsens-300 IMU and JingLing-K50 RTK-GPS, and all these sensors are rigidly mounted on a small vehicle as shown in Fig.~\ref{fig:car}. The last two sequences are from Kaist Urban datasets~\cite{jeong2019complex},
we leverage the IMU measurements and the data from the 3D LiDAR mounted on the vehicle at a tilt of about 45 degrees.
Tab.~\ref{tab:sequence_desci} lists distances and durations of the above mentioned four sequences.
Due to the slow motion of the on-board data, we set the knot distance $\Delta t$ as 0.1 second.
We compute absolute pose error (APE)~\cite{grupp2017evo} with the provided ground truth to compare the CLINS, LIO-SAM and LIOM. Tab.~\ref{tab:accuracy_summary} summarizes RMSE results, and the proposed CLINS shows promising accuracy in the experiments.
Fig.~\ref{fig:map_of_yq01} illustrates the trajectory comparison results with different methods on YQ-01 sequence. Additionally, the bottom figure in Fig.~\ref{fig:map_of_yq01} shows the mapping result of CLINS using YQ-01 sequence. The map, about 700m$\times$500m in size, is consistent with the Google Earth imagery. 

\textbf{Handheld device.} 
Considering that the main advantage of CLINS is non-rigid registration, we conduct qualitative experiments on open source handheld datasets provided from LIO-SAM\footnote{Available at \url{https://drive.google.com/drive/folders/1gJHwfdHCRdjP7vuT556pv8atqrCJPbUq}}. 
Fig.~\ref{fig:mit_map} shows the mapping results for \textit{garden}, \textit{walking} and \textit{small campus} sequences from left to right, respectively. Note that, during collection of \textit{walking} sequence, aggressive motion both in translation and rotation are coupled.

\subsection{Application of Global Continuous Trajectory}
In this section, we introduce an interesting application of global continuous trajectory. 
Generally, a 2D LiDAR-inertial system is challenging to determine 6D pose no matter in the vehicle platform or the handheld platform.
With the assistance of the estimated globally consistent continuous trajectory, 2D LiDAR can provide highly accurate reconstructions.
Fig.~\ref{fig:kaist_map} shows the 3D reconstruction result of Kaist-Urban-07 using 2D LiDAR data from SICK LMS-511. Note that global continuous trajectory is provided by CLINS with 3D LiDAR-inertial system without the participation of 2D LiDAR. 

\subsection{Implementation details}
We adopt the flexible least squares solver Ceres to iteratively solve the NLLS problem and compute derivatives automatically. Typically the non-rigid registration converges within four or five iterations, consuming about 200ms. The computation of jacobians takes most of the computation time and real-time performance can be obtained by using analytical derivatives and exploring the efficient derivative computation for B-Spline~\cite{sommer2020efficient}.

\section{Conclusion}

In this paper, we propose a continuous-time trajectory estimation approach for LiDAR inertial system, termed CLINS. To the best of our knowledge, this is the first open-source continuous time LiDAR-inertial trajectory estimation method. In addiction, we propose a two-state continuous-time trajectory correction method to efficiently and effectively cope with the computationally-intensive loop closure correction.
We compare CLINS against several open-source state-of-the-art discrete-time based algorithms.
The experiments indicate that CLINS outperforms the discrete-time based methods regarding accuracy, which is even more significant in the case of aggressive motion due to the non-rigid registration in our method. 
There are still potential future works to improve the accuracy and efficiency of the system. For example, it is interesting to investigate reducing the number of static control points in the local window.
Instead of automatic derivation, using analytical jacobians and taking full advantage of the recurrence relations~\cite{sommer2020efficient} in the spline computation to reduce computational effort is also worth exploring. 

\section{Acknowledgement}
This work is supported by the National Natural Science Foundation of China under Grant 61836015. We like to thank Hangyu Wu for his assistance with visualizing the results.

\bibliographystyle{IEEEtran}
\bibliography{main.bib}

\begin{thebibliography}{10}
\providecommand{\url}[1]{#1}
\csname url@samestyle\endcsname
\providecommand{\newblock}{\relax}
\providecommand{\bibinfo}[2]{#2}
\providecommand{\BIBentrySTDinterwordspacing}{\spaceskip=0pt\relax}
\providecommand{\BIBentryALTinterwordstretchfactor}{4}
\providecommand{\BIBentryALTinterwordspacing}{\spaceskip=\fontdimen2\font plus
\BIBentryALTinterwordstretchfactor\fontdimen3\font minus
  \fontdimen4\font\relax}
\providecommand{\BIBforeignlanguage}[2]{{%
\expandafter\ifx\csname l@#1\endcsname\relax
\typeout{** WARNING: IEEEtran.bst: No hyphenation pattern has been}%
\typeout{** loaded for the language `#1'. Using the pattern for}%
\typeout{** the default language instead.}%
\else
\language=\csname l@#1\endcsname
\fi
#2}}
\providecommand{\BIBdecl}{\relax}
\BIBdecl

\bibitem{lovegrove2013spline}
S.~Lovegrove, A.~Patron-Perez, and G.~Sibley, ``Spline fusion: A
  continuous-time representation for visual-inertial fusion with application to
  rolling shutter cameras.'' in \emph{BMVC}, vol.~2, no.~5, 2013, p.~8.

\bibitem{ovren2019trajectory}
H.~Ovr{\'e}n and P.-E. Forss{\'e}n, ``Trajectory representation and landmark
  projection for continuous-time structure from motion,'' \emph{The
  International Journal of Robotics Research}, vol.~38, no.~6, pp. 686--701,
  2019.

\bibitem{mueggler2018continuous}
E.~Mueggler, G.~Gallego, H.~Rebecq, and D.~Scaramuzza, ``Continuous-time
  visual-inertial odometry for event cameras,'' \emph{IEEE Transactions on
  Robotics}, vol.~34, no.~6, pp. 1425--1440, 2018.

\bibitem{alismail2014continuous}
H.~Alismail, L.~D. Baker, and B.~Browning, ``Continuous trajectory estimation
  for 3d slam from actuated lidar,'' in \emph{2014 IEEE International
  Conference on Robotics and Automation (ICRA)}.\hskip 1em plus 0.5em minus
  0.4em\relax IEEE, 2014, pp. 6096--6101.

\bibitem{furgale2012continuous}
P.~Furgale, T.~D. Barfoot, and G.~Sibley, ``Continuous-time batch estimation
  using temporal basis functions,'' in \emph{2012 IEEE International Conference
  on Robotics and Automation}.\hskip 1em plus 0.5em minus 0.4em\relax IEEE,
  2012, pp. 2088--2095.

\bibitem{lv2020targetless}
J.~Lv, J.~Xu, K.~Hu, Y.~Liu, and X.~Zuo, ``Targetless calibration of lidar-imu
  system based on continuous-time batch estimation,'' \emph{arXiv preprint
  arXiv:2007.14759}, 2020.

\bibitem{furgale2013unified}
P.~Furgale, J.~Rehder, and R.~Siegwart, ``Unified temporal and spatial
  calibration for multi-sensor systems,'' in \emph{2013 IEEE/RSJ International
  Conference on Intelligent Robots and Systems}.\hskip 1em plus 0.5em minus
  0.4em\relax IEEE, 2013, pp. 1280--1286.

\bibitem{oth2013rolling}
L.~Oth, P.~Furgale, L.~Kneip, and R.~Siegwart, ``Rolling shutter camera
  calibration,'' in \emph{Proceedings of the IEEE Conference on Computer Vision
  and Pattern Recognition}, 2013, pp. 1360--1367.

\bibitem{rehder2014spatio}
J.~Rehder, P.~Beardsley, R.~Siegwart, and P.~Furgale, ``Spatio-temporal laser
  to visual/inertial calibration with applications to hand-held, large scale
  scanning,'' in \emph{2014 IEEE/RSJ International Conference on Intelligent
  Robots and Systems}.\hskip 1em plus 0.5em minus 0.4em\relax IEEE, 2014, pp.
  459--465.

\bibitem{rehder2016general}
J.~Rehder, R.~Siegwart, and P.~Furgale, ``A general approach to spatiotemporal
  calibration in multisensor systems,'' \emph{IEEE Transactions on Robotics},
  vol.~32, no.~2, pp. 383--398, 2016.

\bibitem{furgale2015continuous}
P.~Furgale, C.~H. Tong, T.~D. Barfoot, and G.~Sibley, ``Continuous-time batch
  trajectory estimation using temporal basis functions,'' \emph{The
  International Journal of Robotics Research}, vol.~34, no.~14, pp. 1688--1710,
  2015.

\bibitem{zlot2014efficient}
R.~Zlot and M.~Bosse, ``Efficient large-scale three-dimensional mobile mapping
  for underground mines,'' \emph{Journal of Field Robotics}, vol.~31, no.~5,
  pp. 758--779, 2014.

\bibitem{kaul2016continuous}
L.~Kaul, R.~Zlot, and M.~Bosse, ``Continuous-time three-dimensional mapping for
  micro aerial vehicles with a passively actuated rotating laser scanner,''
  \emph{Journal of Field Robotics}, vol.~33, no.~1, pp. 103--132, 2016.

\bibitem{droeschel2018efficient}
D.~Droeschel and S.~Behnke, ``Efficient continuous-time slam for 3d lidar-based
  online mapping,'' in \emph{2018 IEEE International Conference on Robotics and
  Automation (ICRA)}.\hskip 1em plus 0.5em minus 0.4em\relax IEEE, 2018, pp.
  5000--5007.

\bibitem{park2018elastic}
C.~Park, P.~Moghadam, S.~Kim, A.~Elfes, C.~Fookes, and S.~Sridharan, ``Elastic
  lidar fusion: Dense map-centric continuous-time slam,'' in \emph{2018 IEEE
  International Conference on Robotics and Automation (ICRA)}.\hskip 1em plus
  0.5em minus 0.4em\relax IEEE, 2018, pp. 1206--1213.

\bibitem{haarbach2018survey}
A.~Haarbach, T.~Birdal, and S.~Ilic, ``Survey of higher order rigid body motion
  interpolation methods for keyframe animation and continuous-time trajectory
  estimation,'' in \emph{2018 International Conference on 3D Vision
  (3DV)}.\hskip 1em plus 0.5em minus 0.4em\relax IEEE, 2018, pp. 381--389.

\bibitem{sommer2020efficient}
C.~Sommer, V.~Usenko, D.~Schubert, N.~Demmel, and D.~Cremers, ``Efficient
  derivative computation for cumulative b-splines on lie groups,'' in
  \emph{Proceedings of the IEEE/CVF Conference on Computer Vision and Pattern
  Recognition}, 2020, pp. 11\,148--11\,156.

\bibitem{zhang2014loam}
J.~Zhang and S.~Singh, ``Loam: Lidar odometry and mapping in real-time.'' in
  \emph{Robotics: Science and Systems}, vol.~2, no.~9, 2014.

\bibitem{ceres-solver}
S.~Agarwal, K.~Mierle, and Others, ``Ceres solver,''
  \url{http://ceres-solver.org}.

\bibitem{Mur-Artal2015}
R.~Mur-Artal, J.~M.~M. Montiel, and J.~D. Tardos, ``Orb-slam: a versatile and
  accurate monocular slam system,'' \emph{IEEE Transactions on Robotics},
  vol.~31, no.~5, pp. 1147--1163, 2015.

\bibitem{shan2020lio}
T.~Shan, B.~Englot, D.~Meyers, W.~Wang, C.~Ratti, and D.~Rus, ``Lio-sam:
  Tightly-coupled lidar inertial odometry via smoothing and mapping,''
  \emph{arXiv preprint arXiv:2007.00258}, 2020.

\bibitem{ye2019tightly}
H.~Ye, Y.~Chen, and M.~Liu, ``Tightly coupled 3d lidar inertial odometry and
  mapping,'' in \emph{2019 International Conference on Robotics and Automation
  (ICRA)}.\hskip 1em plus 0.5em minus 0.4em\relax IEEE, 2019, pp. 3144--3150.

\bibitem{jeong2019complex}
J.~Jeong, Y.~Cho, Y.-S. Shin, H.~Roh, and A.~Kim, ``Complex urban dataset with
  multi-level sensors from highly diverse urban environments,'' \emph{The
  International Journal of Robotics Research}, vol.~38, no.~6, pp. 642--657,
  2019.

\bibitem{grupp2017evo}
M.~Grupp, ``evo: Python package for the evaluation of odometry and slam.''
  \url{https://github.com/MichaelGrupp/evo}, 2017.

\end{thebibliography}

\end{document}